\documentclass{article}
\usepackage{ijcai19}

\usepackage{times}
\usepackage{soul}
\usepackage{url}
\usepackage[hidelinks]{hyperref}
\usepackage[utf8]{inputenc}
\usepackage[small]{caption}
\usepackage{graphicx}
\usepackage{amsmath}
\usepackage{booktabs}
\usepackage{algorithm}
\usepackage{algorithmic}
\usepackage{amssymb}
\usepackage{amsmath}
\DeclareMathOperator{\sign}{sign}
\usepackage{array}
\usepackage{makecell}
\usepackage{stfloats} 
\usepackage{mathrsfs}
\usepackage{natbib}

\title{Enhancing the Robustness of Deep Neural Networks \\by Boundary Conditional GAN}

\author{
	Ke Sun$^1$
	\and
	Zhanxing Zhu$^{1,2}$ \footnote{Corresponding author.}\and
	Zhouchen Lin$^{3,4}$
	\affiliations
	$^1$Center for Data Science, Peking University, China\\
	$^2$Beijing Institute of Big Data Research (BIBDR), China\\
	$^3$Key Laboratory of Machine Perception (MOE), School of EECS, Peking University, China\\
	$^4$Cooperative Medianet Innovation Center, Shanghai Jiaotong University, China
	\emails
	\{ajksunke, zhanxing.zhu, zlin\}@pku.edu.cn
}

\begin{document}
	
	\maketitle
	
	\begin{abstract}
		Deep neural networks have been widely deployed in various machine learning tasks. However, recent works have demonstrated that they are vulnerable to adversarial examples: carefully crafted small perturbations to cause misclassification by the network. In this work, we propose a novel defense mechanism called Boundary Conditional GAN to enhance the robustness of deep neural networks against adversarial examples. Boundary Conditional GAN, a modified version of Conditional GAN, can generate boundary samples with true labels near the decision boundary of a pre-trained classifier. These boundary samples are fed to the pre-trained classifier as data augmentation to make the decision boundary more robust. We empirically show that the model improved by our approach consistently defenses against various types of adversarial attacks successfully. Further quantitative investigations about the improvement of robustness and visualization of decision boundaries are also provided to justify the effectiveness of our strategy. This new defense mechanism that uses boundary samples to enhance the robustness of networks opens up a new way to defense adversarial attacks consistently.
	\end{abstract}
	
	\section{Introduction}
	Due to the state-of-the-art performance of deep neural networks, more and more large neural networks are widely adopted in real-world applications. However, recent works~\citep{szegedy2013intriguing,goodfellow6572explaining} have demonstrated that small perturbations are able to fool the networks in producing incorrect prediction by manipulating the input maliciously. The corresponding manipulated samples are called \textit{adversarial examples} that pose a serious threat to the success of deep learning in practice, especially in safety-critical applications.
	
	There exists two types of adversarial attacks proposed in recent literature: white-box attacks and black-box attacks. White-box attacks such as~\citep{szegedy2013intriguing,goodfellow6572explaining,carlini2017towards,akhtar2018threat} allow the attacker to have access to the target model, including architectures and parameters, while under the black-box attacks~\citep{papernot2017practical}, the attacker does not have access to model parameters but can query the oracle, i.e., the targeted DNN, for labels.
	
	Correspondingly, a sizable body of defense strategies are proposed to resist adversarial examples. These defense methods can be mainly categorized into three types:
	
	\paragraph{Adversarial training/Robust optimization.} FGSM adversarial training~\citep{szegedy2013intriguing,goodfellow6572explaining} augmented the training data of classifier with existing types of adversarial examples, usually referred to as first-order adversary. However, some works~\citep{papernot2017practical,tramer2017ensemble,na2017cascade} showed that it is vulnerable to the gradient masking problem. \cite{madry2017towards} studied the adversarial robustness through the lens of robust optimization~\citep{sinha2018certifying} and further trained projected gradient descent~(PGD) adversary as a new form of adversarial training. Note that all of these forms of adversarial training above rely on some specific types of attacks, thus showing relatively good robustness only against the corresponding attacks. However, they may not necessarily defense against other types of attacks consistently.
	
	\paragraph{Input transformation.} Thermometer encoding~\citep{buckman2018thermometer} is a direct input transformation method, employing thermometer encoding to break the linear nature of networks that is stated as the source of adversarial examples~\citep{goodfellow6572explaining}. Defense-GAN~\citep{samangouei2018defense} trained a generative adversarial network (GAN) as a denoiser to project samples onto the data manifold before classifying them. Unfortunately, \cite{athalye2018obfuscated} found that these methods including other input transformations~\citep{ma2018characterizing,guo2017countering} suffered from \textit{obfuscated gradient problem} and can be circumvented by corresponding attacks.
	
	\paragraph{Distillation-type method.} Defensive distillation~\citep{papernot2016distillation} trained the classifier  two rounds using a variant of the distillation~\citep{hinton2015distilling} to learn a smoother network. The approach reduces the model's sensitivity to input variations by decreasing the absolute value of  model's Jacobian matrix and makes it difficult for attackers generate adversarial examples. However, related work~\citep{papernot2017practical} showed that it fails to adequately protect against black-box attacks transferred from other networks ~\citep{carlini2017towards}.  
	
	To design an effective defense method that can be resistant to various types of attacks, we propose a novel defense mechanism called Boundary Conditional GAN that can enhance the robustness of deep neural networks by augmenting boundary samples it generates; and demonstrate its effectiveness against various types of attacks. We leverage the representative power of the state-of-the-art conditional GAN, Auxiliary Classifier GAN~\citep{odena2016conditional} to generate conditional samples. Our key idea is to modify the loss of conditional GAN by additionally minimizing the Kullback-Leibler~(KL) divergence from the predictive distribution to the uniform distribution in order to generate conditional samples near the decision boundary of a pre-trained classifier. These \textit{boundary samples} generated by the modified Conditional GAN are fed to the pre-trained classifier to make the decision boundary more robust. The crucial point why Boundary Conditional GAN can  help to consistently defense a wide range of attacks rather than a specific type of attacks is that the boundary samples might represent almost all the potential directions of constructed adversarial examples. 
	
	We empirically show that the new robust model can resist various types of adversarial examples and exhibits consistent robustness to these attacks compared with FGSM, PGD adversarial training, defensive distillation and Defense-GAN. Furthermore, we quantify the enhancement of robustness on MNIST, Fashion-MNIST and CIFAR10 datasets. Finally, we visualize the change of decision boundaries to further demonstrate the effectiveness of our approach. Boundary Conditional GAN supplies us a new way to design a consistent defense mechanism against both existing and future attacks.
	
	\section{Preliminaries}
	Before introducing our approach, we firstly present some preliminary knowledge about different adversarial attacks employed in this work and necessary background information about Conditional GANs.
	\subsection{Adversarial Attacks}
	Adversarial attacks aim to find a small perturbation $\eta$, usually constrained by $l_{\infty}$-norm, and then add the perturbation to a legitimate input $x \in  \mathbb{R}^n$ to craft adversarial examples $ \widetilde{x}=x+\eta$ that can fool the deep neural networks. In this paper, we consider white-box attacks, where the adversary has full access to the neural network classifier~(architectures and weights).\\
	\textbf{Fast Gradient Sign Method}~(\textbf{FGSM}, \cite{goodfellow6572explaining}) is a simple but effective attack for an $l_{\infty}$-bounded adversary and an adversarial example can be obtained by:
	\begin{align}
	\tilde{x} = x + \epsilon \sign (\nabla _{x}J(x, y)),
	\end{align}
	where $\tilde{x}$ denotes the adversarial example crafted from an input $x$ and $\epsilon$ measures the magnitude of the perturbation. $J(x,y)$ denotes the loss function of the classifier given the input $x$ and its true label $y$, e.g., the cross entropy loss. FGSM is widely used in attacks and design of defense mechanisms.\\
	\textbf{Projected Gradient Descent attack}~(\textbf{PGD}, ~\cite{madry2017towards}) is an iterative attack method and can be regarded as a multi-step variant of FGSM:
	\begin{align}
	x^{t+1}=\text{proj}_{\Omega}(x^t + \alpha \sign(\nabla _{x}J(x, y))),
	\end{align}
	where $\Omega\!=\![0,255]^{n} \,\bigcap \, \{\tilde{x}\, | \, \Vert \tilde{x}-x \Vert_{\infty} \le \epsilon \}$ and $\alpha$ is the step size. \cite{madry2017towards} showed that (the $l_{\infty}$ version of) PGD is equivalent to Basic Iterative Method (BIM), another important iterative attacks. In this paper, we use PGD attack to represent a variety of iterative attacks.\\
	\textbf{Carlini-Wagner~(CW) attack}~\citep{carlini2017towards} is an effective optimization-based attack. In many cases, it can reduce the classifier accuracy to almost zero. The perturbation $\eta$ is found by solving an optimization of the form:
	\begin{equation} \begin{aligned} 
	&\mathop{\min}_{\eta \in  \mathbb{R}^n} \Vert \eta \Vert_{p} + c \cdot f(x+\eta)\\
	&\ \ s.t. \ \ \ \  x + \eta \in [0,1]^n,
	\end{aligned} \end{equation} 
	where $\Vert\!\cdot\!\Vert_p$ is the $l_p$ norm, $f$ is the objective function that is defined to drive the example $x$ to be misclassified, e.g., $f(x^{\prime})=(\mathop{\max}\limits_{i\neq t}(F(x^{\prime})_i)-F(x^{\prime})_t)^{+}$ where $t$ denotes a different class and $F$ is the classifier. $c$ represents a suitably chosen constant. Although various norms could be considered such as the $l_0,l_2,l_{\infty}$ norms, we choose CW attack with $l_2$-norm due to the convenience of computation.
	
	\subsection{Conditional GANs}
	\textbf{Generative Adversarial Networks}~(\textbf{GANs}, \cite{goodfellow2014generative}) consist of two neural networks trained in opposition to one another. The generator $G: \mathbb{R}^{k} \to \mathbb{R}^{n}$ maps a low-dimensional latent space to the high dimensional sample space of $x$. The discriminator $D: \mathbb{R}^{n} \to [0,1]$ is a binary classifier, discriminating the real and fake inputs generated by the generator $G$. The generator and discriminator are trained in an alternating fashion to minimize the following min-max loss:
	\begin{align}
	\mathop{\min}_{G} \mathop{\max}_{D} \mathbb{E}_{x\sim p_{\text{data}}(x)}[\log D(x)] + \mathbb{E}_{z\sim p_z(z)}[\log (1-D(G(z)))],
	\end{align}	
	where $z$ is the noise, usually following a simple distribution $p(z)$, such as Gaussion distribution. The objective functions of discriminator $D$ and generator $G$ are as follows:
	\begin{equation} \begin{aligned} 
	&L_{D} = \mathbb{E}_{x}[\log D(x)]+ \mathbb{E}_{z}[\log (1-D(G(z)))],\\
	&L_{G} = \mathbb{E}_{z}[\log D(G(z))].
	\end{aligned} \end{equation} 
	\textbf{Auxiliary Classifier GAN}~(\textbf{ACGAN}, ~\cite{odena2016conditional}) leverages both the noise $z$ and the class label $c$ to generate each sample from the generator $G$, $X_{fake}=G(c,z)$. The discriminator gives a probability distribution $P(S|X)$ over sources, i.e., real or fake examples, and a probability $P(C|X)$ over the class labels respectively, $P(S|X), P(C|X)=D(X)$, where $S$ denotes the sources and $C$ denotes the class labels. The objective function has two parts: the log-likelihood $L_{S}$ of the correct source and the log-likelihood $L_{C}$ of the correct class.
	\begin{equation} \begin{aligned} 
	&L_{S} = \mathbb{E}_x[\log P(S=real|X_{real})]+\\
	& \qquad \qquad \qquad \qquad \qquad \qquad \mathbb{E}_z[\log P(S=fake|X_{fake})],\\
	&L_{C} = \mathbb{E}_x[\log P(C=c|X_{real})]+ \mathbb{E}_z[\log P(C=c|X_{fake})].
	\end{aligned}  \label{eq:acgan}
	\end{equation} 
	\noindent $D$ is trained to maximize $L_{S}+L_{C}$ while $G$ is trained to maximize $L_{C}-L_{S}$. ACGAN learns a representation from $z$ that is independent of class label and we make a choice to use ACGAN considering its state-of-the-art performance.
	
	\section{The Proposed Boundary Conditional GAN}
	In this section, we will elaborate our approach \textit{Boundary Conditional GAN}. First, we will detail our motivation why we consider to use boundary samples to defense against adversarial examples. Then, to verify that the proposed Boundary Conditional GAN can generate boundary samples near the decision boundary, we implement an experiment in a 2D classification task, in which we visualize the decision boundary and demonstrate that adding the Kullback-Leibler~(KL) penalty to the loss of conditional GAN can force the generated samples with labels to be near the decision boundary of original classifier. Finally, we introduce the procedure how to use Boundary conditional GAN to generate boundary samples to enhance the robustness of our pretrained classifier.
	
	\subsection{Motivation}
	\begin{figure}[t]
		\centering
		\centering\includegraphics[scale=0.75]{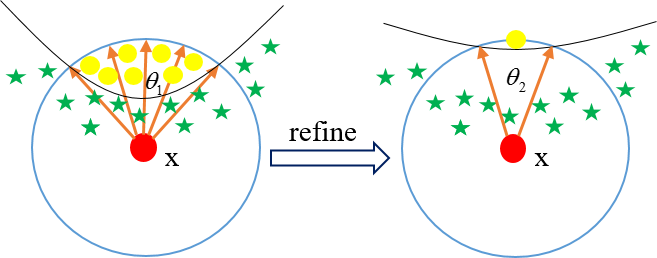}
		\caption{The effect of boundary samples against adversarial examples. The red point is a clean example $x$ and yellow points represent the existing adversarial examples. Green stars denote the boundary samples generated by Boundary Conditional GAN. And the black solid curves represent the decision boundary before and after considering the boundary examples.}
		\label{motivation}
	\end{figure}
	
	From the perspective of attacks, the optimization-based adversarial attacks, e.g., FGSM and PGD, solve the following optimization problem to different extents so that they obtain different approximations of the optimal adversarial attack.
	\begin{equation} \begin{aligned} 
	&\mathop{\min}_{\eta} \ \ \Vert \eta \Vert \\
	&\ s.t. \ F(x+\eta) \neq F(x),
	\end{aligned} \end{equation}
	where $F$ denotes the classifier. However, adversarial attacks exist in many different directions around input $x$ \citep{goodfellow2018making} and the constructed attacks above only represent some potential directions of them, which also explains why adversarial training based on these attacks has limited defense power against other types of attacks. Considering a clean example $x$ with a $l_{\infty}$ norm ball near the current decision boundary represented by black solid curve in the left part of Figure~\ref{motivation}, we can observe that there exist adversarial examples, i.e., the coverage of yellow points, \emph{in some continuous directions of an angle $\theta_1$ given the magnitude of perturbation $\| \eta\| $}. Samples near the decision boundary, i.e., the green stars, can represent almost all directions of adversarial examples, thereby it is natural to consider to use boundary samples to refine the decision boundary by data augmentation to help the classifier defense against various types of adversarial attacks. It is expected that the decision boundary could be refined from left panel to right one in Figure~\ref{motivation} through considering boundary samples, thus shrinking the coverage of potential adversarial examples with an angle of direction from $\theta_1$ to $\theta_2$. Consequently, the improved decision boundary reduces the number of adversarial examples significantly.
	
	From the perspective of defense, the consistent effectiveness against a variety of attacks is of vital importance. \cite{athalye2018obfuscated} pointed out that a strong defense should be robust not only against existing attacks, but also against future attacks. The key point of our motivation is that boundary samples could represent almost all potential directions of adversarial attacks and might exhibit consistent robustness against various types of attacks, which takes both aspects of attacks and defense into consideration. 	
	
	\subsection{Boundary Conditional GAN}
	
	\begin{figure}[b]
		\centering
		\centering\includegraphics[width=.45\textwidth,trim=150 50 120 20,clip]{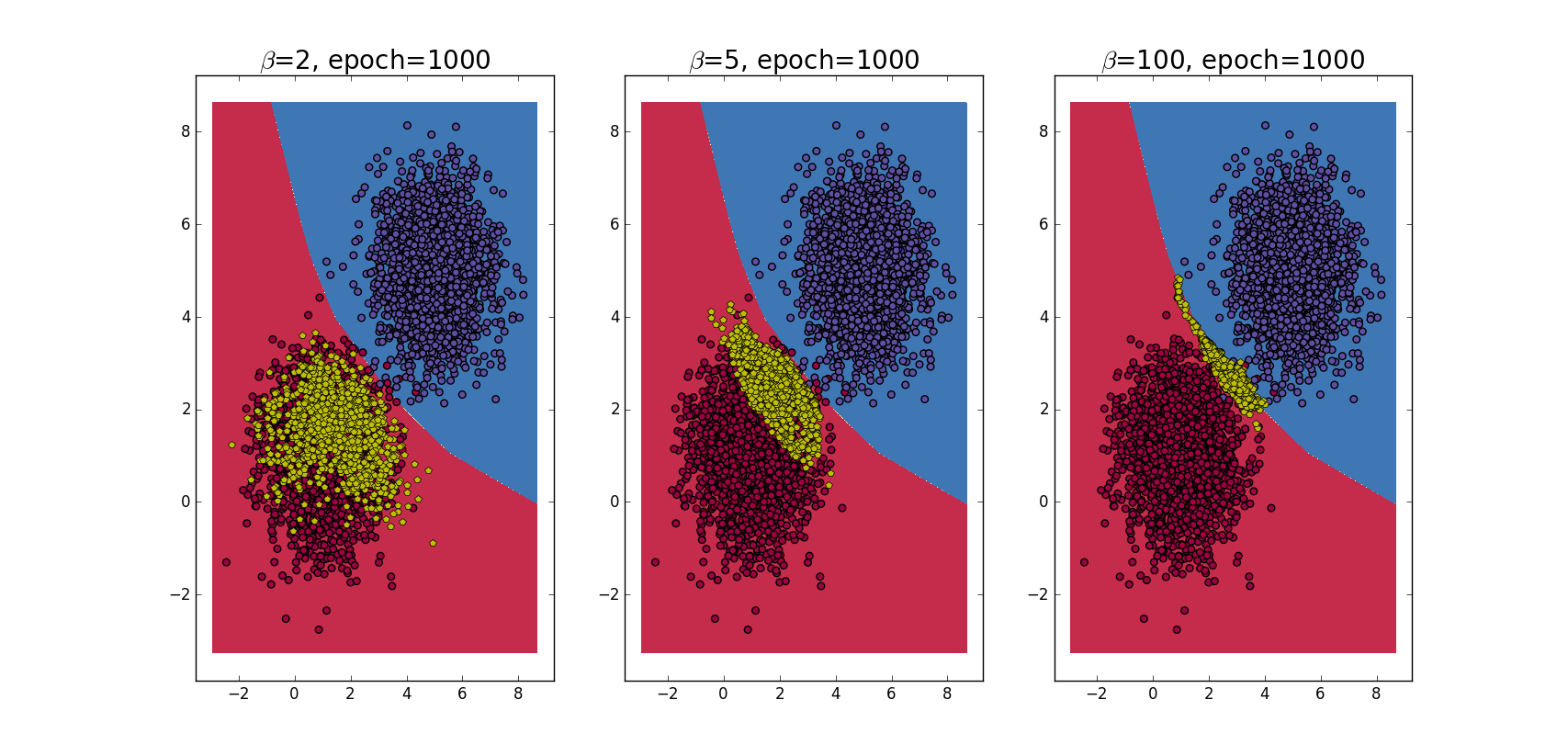}
		\caption{The generated boundary samples in different $\beta$ from proposed Conditional GAN. Red and blue points denote the data of different classes and yellow points represent the generated boundary samples.}
		\label{fig:bcgan_boundary}
	\end{figure}
	
	It is highly intuitive that the predictive distribution by the classifier for the samples near the decision boundary is close to a uniform distribution due to the ambiguity which class the boundary samples belong to. For example, in a classification task with two groups, the classification probabilities of samples on the decision boundary is a vector $[0.5, 0.5]$.  
	Therefore, in order to facilitate the conditional GAN to generate more samples near the decision boundary of original classifier, we propose to add an additional penalty to force the predictive distribution of generated samples through the original classifier to be a uniform distribution. 
	The  new generator $G$ loss~\citep{lee2017training} is as follows,
	\begin{equation} \begin{aligned} 
	L_{G_{KL}} = \underbrace{L_{G}}_{(a)} + \underbrace{\beta \mathbb{E}_{P_{G}(x)}[KL(U(y)||P_{\theta}(y|x))]}_{(b)}, \label{eq:bcgan}
	\end{aligned} \end{equation}
	where $L_{G}$ is the original generator loss of conditional GAN mentioned in (6) and (7), e.g., for ACGAN, $L_G = L_{C}-L_{S}$. $\theta$ are parameters of the original classifier rather than auxiliary classifier in ACGAN, which are fixed during the training of the proposed Conditional GAN. $U(\cdot)$ is the uniform distribution and $\beta > 0$ is a penalty parameter. 
	The first term (a) corresponds to the original conditional GAN loss since we would like to guarantee that the generated samples are near the original distribution of the corresponding class and not too far from the data manifold. The KL divergence term (b) forces the generator to generate samples whose predictive distribution through the original classifier is close to the uniform one, i.e., samples near the decision boundary of the pretrained classifier, by minimizing the KL loss while training.
	
	To illustrate the effectiveness of the new GAN, we firstly implement an experiment on the modified conditional GAN in a 2D classification task; and show that the new GAN loss can help the conditional GAN generate samples near the decision boundary with corresponding labels. The training data are simulated from two 2D Gaussian distributions in red and blue respectively shown in Figure~\ref{fig:bcgan_boundary}. For both the generator and the discriminator, we use a fully-connected neural network with 3 hidden layers. We visualize the decision boundary and samples generated by the proposed boundary conditional GAN in Figure~\ref{fig:bcgan_boundary}. It shows that the new KL penalty can indeed generate conditional samples in yellow near the decision boundary of original classifier. And generated samples in yellow with different $\beta$ are close to decision boundary to different extents.
	
	\subsection{Defense Mechanism}	
	In practice, we can easily access a pre-trained classifier for a specified machine learning task and then design a defense mechanism based on that. Due to the influence of the new loss Eq.~\eqref{eq:bcgan}, there exists a slight decreasing of precision for the obtained conditional GAN by directly training the modified conditional GAN from scratch. To overcome this issue, we inject clean examples during the data augmentation to maintain the accuracy of original classifier.
	
	\begin{figure}[b]
		\centering
		\centering\includegraphics[scale=0.35]{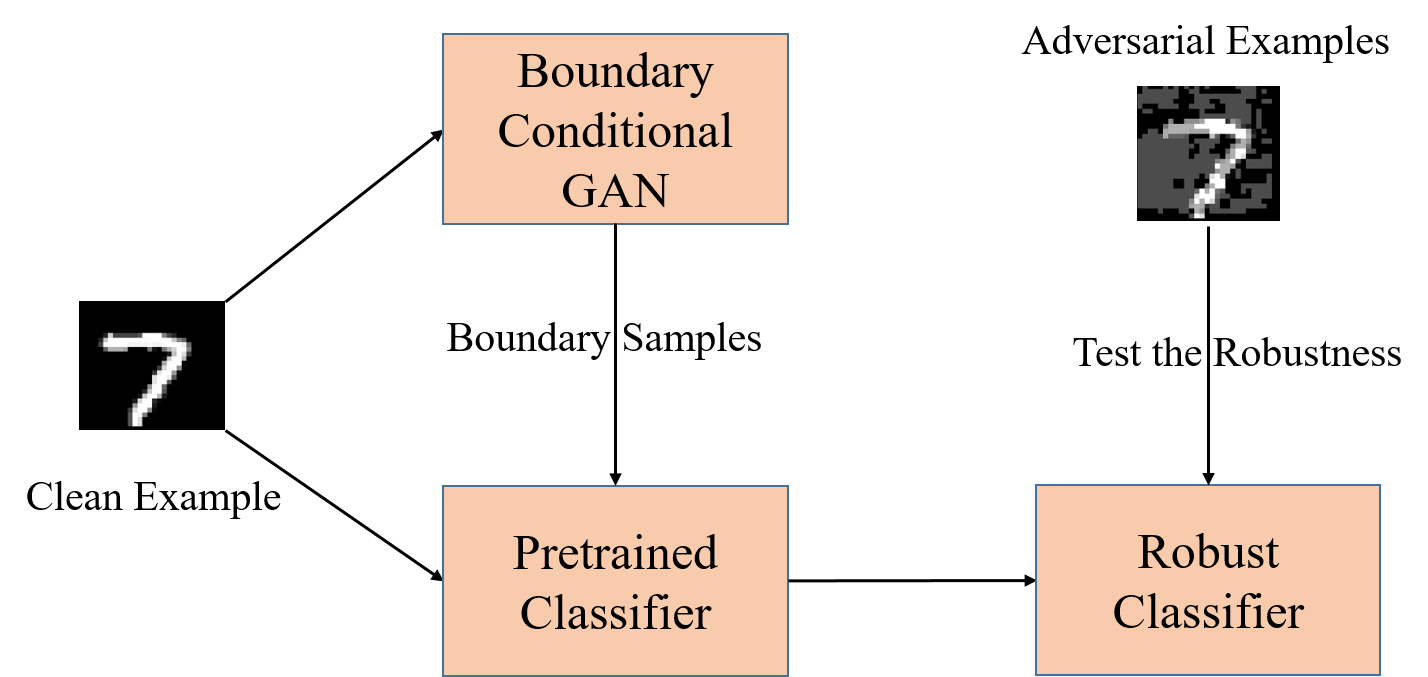}
		\caption{Flow chart of our defense mechanism against adversarial examples.}
		\label{fig:flow_chart}
	\end{figure}
	
	Here, we describe our procedure of defense mechanism as follows and corresponding flow chart is shown in Figure~\ref{fig:flow_chart}.
	\begin{enumerate}
		\item Pre-train a classifier, i.e., the target model to defense, on the specified dataset;
		\item Train the modified conditional GAN with the new KL loss; Eq.~\eqref{eq:bcgan}, forcing the conditional GAN to generate boundary samples;
		\item Feed the boundary samples with corresponding labels to the pre-trained classifier to refine the decision boundary by data augmentation;
		\item Evaluate the robustness of the final classifier against various types of adversarial attacks.
	\end{enumerate}
	
	\section{Experiments}
	\label{sec:exp}
	In this section, in order to demonstrate the effectiveness by our approach on improving the robustness of deep neural networks, we conduct experiments on MNIST, Fashion-MNIST and CIFAR10 datasets from three aspects as follows.
	
	\paragraph{Defense against adversarial attacks.} We empirically show that the new robust model by Boundary Conditional GAN can resist various adversarial attacks, e.g., FGSM, PGD and CW attacks. We compare the result with Defensive Distillation~\citep{papernot2016distillation}, Defense-GAN~\citep{samangouei2018defense}, a defense approach also based on GAN, and FGSM adversarial training~\citep{szegedy2013intriguing,goodfellow6572explaining} and PGD adversarial training~\citep{madry2017towards} which are regarded as commonly accepted baselines of defense. Consistent robustness can be observed through our detailed analysis.
	
	\paragraph{Quantitative analysis of robustness.} To quantify the enhancement of robustness by our Boundary Conditional GAN, we quantitatively evaluate the robustness on MNIST, Fashion-MNIST and CIFAR10 and compare that with other defensive approaches. 
	
	\paragraph{Visualization of decision boundaries.} To verify the improvement on robustness of decision boundaries, we visualize the change of decision boundaries around the input $x$.
	
	\subsection{Defense against Adversarial Attacks}
	We test the robustness of the original and improved classifier by Boundary Conditional GAN against various attack strategies compared with FGSM adversarial training~\citep{szegedy2013intriguing,goodfellow6572explaining}, PGD adversarial training~\citep{madry2017towards}, Defensive Distillation~\citep{papernot2016distillation} and Defense-GAN~\citep{samangouei2018defense}.
	
	\noindent \textbf{Settings of various attack strategies.}
	We present the experimental results by using three different strategies: FGSM, PGD and CW. We perform FGSM with different magnitude $\epsilon$ and PGD attack for 40 iterations of projected GD on MNIST, Fashion-MNIST and 8 iterations on CIFAR10. Next, we perform $l_2$-norm CW attack with 1,000 test samples. 
	
	\noindent \textbf{Settings of baselines.}
	FGSM and PGD adversarial training  are trained with adversarial examples generated by standard FGSM and PGD attacks mentioned above with different magnitude. Defensive distillation is trained with soft labels under Temperature $T=100$, just the same as original paper~\citep{papernot2016distillation}. Defense-GAN, as another baseline, is trained with WGAN \citep{arjovsky2017wasserstein} first and has $L=12000$ at inference time on all datasets.
	
	\noindent \textbf{Settings of architectures of deep neural networks.}
	The classifier on MNIST and Fashion-MNIST has two convolutional layers and one fully-connected layer. For CIFAR10, we directly leverage ResNet18. Meanwhile, the architecture of conditional GAN, i.e., ACGAN, is adopted by the original one~\citep{odena2016conditional} so that the state-of-the-art performance can be maintained. 
	
	We train ACGAN on the new GAN loss Eq.~\eqref{eq:bcgan} on the corresponding dataset to generate boundary samples and then feed these boundary samples to refine the classifier by data augmentation. Finally, we leverage the enhanced classifier to test the effectiveness of robustness against various types of adversarial attacks in Table~\ref{tab:acc}.
	
	\begin{table}[htbp]
		\centering
		\caption{Classification accuracies (\%) using different defense strategies against various attacks on the MNIST, Fashion-MNIST and CIFAR10. ``BCGAN'' denotes the Boundary Conditional GAN and ``Adv.Tr'' is adversarial training. Defense-GAN is with $L$=12000. The number in bracket after FGSM is the perturbation magnitude $\epsilon$.}
		\label{tab:acc}
		\begin{center}
			\centering\textbf{MNIST}
			\begin{tabular}{c|c|c|c|c|c}
				\hline\hline
				\bf Defense&Clean&\makecell{FGSM\\(0.2)}&\makecell{FGSM\\(0.4)}&PGD&\makecell{CW\\$l_2$}\\
				\hline  
				Original                                &99.2&74.5&31.3&11.0&32.6\\
				\makecell{FGSM Adv.Tr\\($\epsilon$=0.2)}&\bf99.3&96.2&84.7&73.3&85.2\\
				\makecell{FGSM Adv.Tr\\($\epsilon$=0.4)}&98.8&96.3&92.3&44.5&96.3\\
				PGD Adv.Tr                              &99.1&97.1&92.6&93.6&96.4\\
				Distillation                            &96.9&96.0&96.0&96.0&2.1\\
				Defense-GAN                             &90.1&83.1&71.2&77.0&34.4\\
				\hline 
				BCGAN                                   &97.7&\bf97.4&\bf97.3&\bf97.4&\bf97.5\\
				\hline\hline
			\end{tabular}
		\end{center}
		
		\centering \textbf{Fashion-MNIST}
		\begin{tabular}{c|c|c|c|c|c}
			\hline\hline
			\bf Defense&Clean&\makecell{FGSM\\(0.05)}&\makecell{FGSM\\(0.15)}&PGD&\makecell{CW\\$l_2$}\\
			\hline  
			Original                                 &\bf90.9&60.7&18.6&0.0&0.0\\
			\makecell{FGSM Adv.Tr\\($\epsilon$=0.05)}&90.3&81.1&58.5& 2.0&54.4\\
			\makecell{FGSM Adv.Tr\\($\epsilon$=0.15)}&89.0&80.5&76.0&14.9&82.3\\
			PGD Adv.Tr                               &83.5&81.5&77.7&67.3&84.9\\
			Distillation                             &88.1&80.4&80.2&80.1&0.0\\
			Defense-GAN                              &83.5&78.9&69.0&49.8&37.9\\
			\hline 
			BCGAN                                    &85.4&\bf83.8&\bf83.8&\bf83.6&\bf85.9\\
			\hline\hline
		\end{tabular}
		
		\begin{center}
			\centering \textbf{CIFAR10}
			\begin{tabular}{c|c|c|c|c|c}
				\hline\hline
				\bf Defense&Clean&\makecell{FGSM\\(0.01)}&\makecell{FGSM\\(0.03)}&PGD&\makecell{CW\\$l_2$}\\
				\hline  
				Original                                 &81.7&44.9&23.0&1.6&0.1\\
				\makecell{FGSM Adv.Tr\\($\epsilon$=0.01)}&81.0&67.9&46.1&25.4&6.3\\
				\makecell{FGSM Adv.Tr\\($\epsilon$=0.03)}&78.5&69.1&53.0&35.5&21.6\\
				PGD Adv.Tr                               &76.2&68.3&54.3&38.9&34.4\\
				Distillation                             &\bf84.9&75.7&75.5&75.1&0.0\\
				Defense-GAN                              &44.0&43.1&42.3&41.8&42.2\\
				\hline 
				BCGAN                                    &80.9&\bf80.3&\bf80.2&\bf80.2&\bf80.7\\
				\hline\hline
			\end{tabular}
		\end{center}
	\end{table}
	
	Table~\ref{tab:acc} shows the classification accuracies under different defense strategies across various attacks on all the three datasets. An important observation is that the Boundary Conditional GAN significantly outperforms Defensive Distillation, Defense-GAN and FGSM, PGD adversarial training with different magnitude $\epsilon$ against all attacks especially on CIFAR10. Concretely speaking, adversarial training with stronger attacks, e.g., larger $\epsilon$ for FGSM or PGD, exhibits better robustness but they more easily suffer from overfitting to the crafted adversarial examples, showing a larger drop on clean accuracy. However, these types of adversarial training perform worse than other defensive methods on larger datasets such as CIFAR10. In addition, Defensive Distillation is on par with the state-of-the-art performance of BCGAN across gradient-based attacks but it fails to defense stronger CW attack, which is also demonstrated in ~\citep{carlini2017towards}. We re-implement Defense-GAN based on original paper~\citep{samangouei2018defense} due to the different setting. However, the pratical difficulities especially the choice of hyper-parameters of Defense-GAN, which is also discussed in \citep{samangouei2018defense}, hinder the effectiveness of this method, resulting its limited defensive performance especially on CIFAR10, in which the original paper of Defense-GAN~\citep{samangouei2018defense} has not provided corresponding experimental result. For the Boundary Conditional GAN, the consistent robustness of our method is easy to observe although it slightly decreases the accuracy on clean data due to the influence of limited accuracy of conditional GAN, i.e., ACGAN.
	
	
	%
	
	\subsection{Quantitative Analysis of Robustness}
	In order to further demonstrate the enhancement of robustness against adversarial attacks, we quantitatively investigate the enhancement of robustness. 
	
	\begin{table}[b]
		\centering
		\caption{Enhancement of robustness by Boundary conditional GAN in comparison between the original model and the robust model on MNIST, Fashion-MNIST and CIFAR10. ``Adv.Tr1'' and ``Adv.Tr2'' denotes adversarial training with different $\epsilon$, which are the same as the setting in Table~\ref{tab:acc}.}
		\label{tab:robustness}
		\begin{center}	
			\begin{tabular}{c|c|c|c}
				\hline
				\bf Robustness &\bf MNIST &\bf F-MNIST &\bf CIFAR10 \\
				\hline  
				Original       &0.40    &0.15    &0.16\\
				FGSM Adv.Tr1   &0.77    &0.27    &0.22\\
				FGSM Adv.Tr2   &0.86    &0.47    &0.33\\
				PGD  Adv.Tr    &0.88    &0.58    &0.41\\
				Distillation   &1.67    &1.65    &1.84\\
				Defense-GAN    &0.10    &0.11    &0.12\\
				\hline
				BCGAN          &\bf 1.94&\bf 1.70&\bf 2.00\\
				\hline
			\end{tabular}
		\end{center}
	\end{table}	
	
	\begin{figure*}[ht]
		\centering
		\centering\includegraphics[width=.8\textwidth,trim=50 350 50 350,clip]{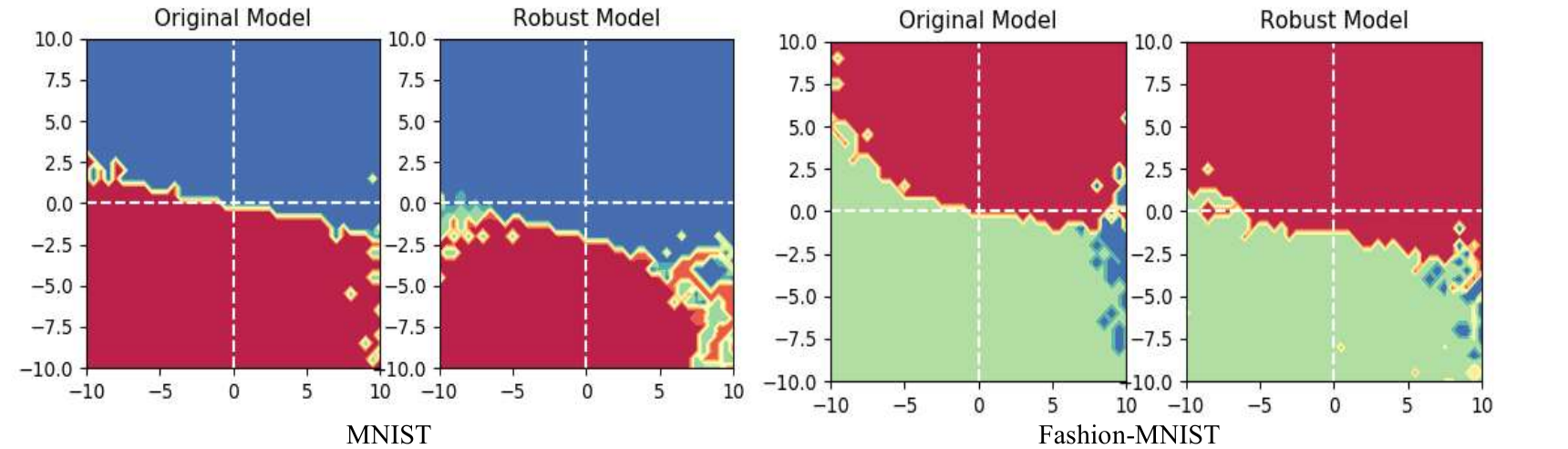}
		\caption{Change of decision boundaries of original model and robust model by Boundary Conditional GAN on MNIST and Fashion-MNIST.}
		\label{fig:vis}
	\end{figure*}
	
	We extend the measure of robustness~\citep{papernot2017practical} to the adversarial behavior of source-target class pair misclassification within the context of classifiers built using DNNs. The robustness of a trained DNN model $F$ is:
	\begin{equation} \begin{aligned} 
	&\rho_{adv}(F)=\mathbb{E}_{\mu}[\Delta_{adv}(X,F)],
	\end{aligned} \end{equation}
	where inputs $X$ are drawn from data distribution $\mu$, and $\Delta_{adv}(X,F)$ is defined to be the minimum perturbation required to misclassify sample $x$ in each of the other classes. We reformulate $\Delta_{adv}(X,F)$ for simplicity as follows:
	\begin{equation} \begin{aligned} 
	&\Delta_{adv}(X,F) = \mathop{\arg \min}_{\eta} \{\Vert \eta \Vert : F(X+\eta) \neq F(X) \}
	\end{aligned} \end{equation}
	
	The higher the average minimum perturbation required to misclassify a sample is, the more robust a DNN is against adversarial examples. Then, we evaluate whether Boundary Conditional GAN increases the robustness metric on the three datasets. Unlike the original method~\citep{papernot2017practical}, we do not approximate the metric but search all perturbations for each sample $x$ with certain precision.
	
	As shown in Table~\ref{tab:robustness}, Boundary Conditional GAN significantly improves the robustness of deep neural network on the three datasets. More importantly, the enhancement of robustness by Boundary Conditional GAN exceeds various types of adversarial training and Defense-GAN dramatically, showing the state-of-the-art performance. It is interesting to find that the real robustness of Defense-GAN is poor and the underlying reason might lie in the \textit{mode collapse} problem mentioned in the NIPS 2016 GAN tutorial~\citep{goodfellow2016nips}. Modified conditional GAN applied in our approach might partially avoid this problem due to the good property of conditional GAN.
	
	\subsection{Visualization of Decision Boundaries}
	\noindent In this part, we visualize the effect of boundary samples by comparing the change of decision boundary on two given directions in Figure~\ref{fig:vis}. We apply the visualization method proposed by~\cite{wu2018understanding}, in which the two directions of axes are chosen as follows. 
	
	Denoting the gradient $g(x):=\nabla_x J(x)$, the first direction is selected as the locally averaged gradient, 
	\begin{equation} \begin{aligned} 
	G(x)=\mathbb{E}_{\xi \sim \mathcal N(0,\sigma^2)}[g(x+\xi)], \label{eq:G}
	\end{aligned} \end{equation}
	where $G$ denotes the smoothed gradient. The motivation using this smoothed gradient is the shattered gradient phenomenon studied in \citep{balduzzi2017shattered}, observing that the gradient $g(x):=\nabla_x J(x)$  is very noisy; and one way to alleviate it is to smooth the landscape $J$, thereby yielding a more informative direction than $g$. We choose $\sigma$ = 1, $m = 1000$ and the expectation in \eqref{eq:G} is estimated by $\frac{1}{m}\sum_{i=1}^{m}g(x+\xi_i)$. As shown in Figure~\ref{fig:vis}, the horizontal axis represents the direction of smoothed gradient $G$ and the vertical axis denotes the orthogonal direction $h:=g-\langle g,\hat{G} \rangle \hat{G}$. Each point in the 2-D plane corresponds to an image perturbed by $u$ and $v$ along each direction,
	$$\text{clip}(x+u \hat{g}+v\hat{h}, 0, 255),$$
	where the origin $x$ denotes the considered clean example, i.e. the crossover point of the two dashed axes. The different colors represent the different classes of the perturbed images in the direction $u$ and $v$. The left part of the subfigure for each dataset depicts the decision boundary of the original model around the clean image, while the right part denotes that of the robust model achieved by Boundary Conditional GAN.
	
	We visualize the improvement of decision boundaries on MNIST and Fashion-MNIST. As shown on the left part for MNIST in Figure~\ref{fig:vis}, the region in blue above the central point has been enlarged from the original model~(left) to the robust model~(right), indicating that only larger perturbations could attack the new model successfully. The similar situation can also be observed on the right~(Fashion-MNIST), where the red region has been expanded around the decision boundary, exhibiting the sufficient robustness of our approach. All of the results in Figure~\ref{fig:vis} suggest that the robustness of the classifier has been improved by Boundary Conditional GAN.
	
	\section{Discussions and Conclusion}
	Through our empirical observation, we found that
	the diversity and accuracy of conditional GAN is of significant importance for our defense mechanism. The more diversity of the generated samples by conditional GAN have, the more improvement of robustness can be observed. Furthermore, the more accurate of conditional GAN is, the less decreasing accuracy caused by misclassified samples generated by conditional GAN can be obtained. Moreover, other strategies that can generate samples near the decision boundary can also be leveraged to design defense mechanisms.
	
	In this work, we have proposed a novel defense mechanism called Boundary Conditional GAN to enhance the robustness of decision boundary against adversarial attacks. We leverage the modified Conditional GAN by additionally minimizing the Kullback-Leibler~(KL) divergence from the predictive distribution to the uniform distribution in order to generate samples near the decision boundary of the pre-trained classifier. These boundary example might capture the different directions of various adversarial attacks. Then we feed the boundary samples to the pre-trained classifier to refine the decision boundary. We empirically show that the new robust model can be resistant to various types of adversarial attacks and quantitatively evaluation on the enhancement of robustness and visualization of the improvement of decision boundaries are also provided.
	
	In summary, leveraging boundary samples by Boundary Conditional GAN opens up a new way to design defense mechanism against various types of adversarial examples consistently in the future.
	
	\bibliographystyle{named}
	\bibliography{ijcai19}
	
\end{document}